\documentclass[conference]{IEEEtran}
\usepackage{times}

\usepackage[numbers]{natbib}
\usepackage{notoccite}
\usepackage{multicol}
\usepackage[bookmarks=true]{hyperref}
\usepackage{graphicx}
\usepackage{braket}
\usepackage{textcomp}
\usepackage{xcolor}
\usepackage{subcaption}
\usepackage{float}
\usepackage{multicol}
\usepackage{caption}
\usepackage{soul}
\usepackage{multirow}
\usepackage{array}
\usepackage{makecell}
\usepackage{hyperref}
\usepackage{url}
\usepackage{amssymb}
\usepackage{amsmath}
\usepackage{amsfonts}

\usepackage[T1]{fontenc}
\usepackage[utf8]{inputenc}
\usepackage{babel}
\usepackage[font=small,labelfont=bf]{caption}

\newcommand{\name}{\textsc{DeltaHands}}
\newcommand{\namespace}{\textsc{DeltaHands}\space}

\begin{document}

\title{\name: A Synergistic Dexterous Hand Framework Based on Delta Robots}


\author{
Zilin Si$^{1}$, Kevin Zhang$^{1}$, Oliver Kroemer$^{1}$, F. Zeynep Temel$^{1}$
\thanks{ $^{1}$Carnegie Mellon University Robotics Institute, PA, USA 
        \texttt{\{zsi, klz1, okroemer, ztemel\}@andrew.cmu.edu}
        
       }
}


\maketitle

\begin{abstract}
Dexterous robotic manipulation in unstructured environments can aid in everyday tasks such as cleaning and caretaking. Anthropomorphic robotic hands are highly dexterous and theoretically well-suited for working in human domains, but their complex designs and dynamics often make them difficult to control. By contrast, parallel-jaw grippers are easy to control and are used extensively in industrial applications, but they lack the dexterity for various kinds of grasps and in-hand manipulations. In this work, we present \name, a synergistic dexterous hand framework with Delta robots. The \namespace are soft, easy to reconfigure, simple to manufacture with low-cost off-the-shelf materials, and possess high degrees of freedom that can be easily controlled. By leveraging hand synergies, \name' dexterity can be adjusted for different applications with further reduced control complexity. We characterize the Delta robots' kinematics accuracy, force profiles, and workspace range to assist with hand design. Finally, we evaluate the versatility of \namespace by grasping a diverse set of objects and by using teleoperation to complete three dexterous manipulation tasks: cloth folding, cap opening, and cable arrangement. We open-source our hand framework at \href{https://sites.google.com/view/deltahands/}{https://sites.google.com/view/deltahands/}.

\end{abstract}

\IEEEpeerreviewmaketitle

\section{Introduction}
As robots advance closer to assisting humans at home, the design of their end-effectors becomes crucial in ensuring safety and functionality for complex household manipulation tasks.
While parallel jaw grippers are widely used for industrial tasks due to their simple one degree of freedom (DoF) control, many household tasks require hands that are dexterous and can safely adapt to various objects.
Anthropomorphic robotic hands~\cite{deimel2013compliant, deimel2016novel, andrychowicz2020learning, puhlmann2022rbo} possess dexterous fingers capable of accomplishing complex in-hand manipulation tasks; however, they tend to be difficult to control, bulky, and costly to manufacture. We aim to introduce a framework for developing hands that are easy to configure, build, and control for dexterous manipulation.

\begin{figure}    \centering
    \vspace{1mm}
    \includegraphics[width=0.98\linewidth]{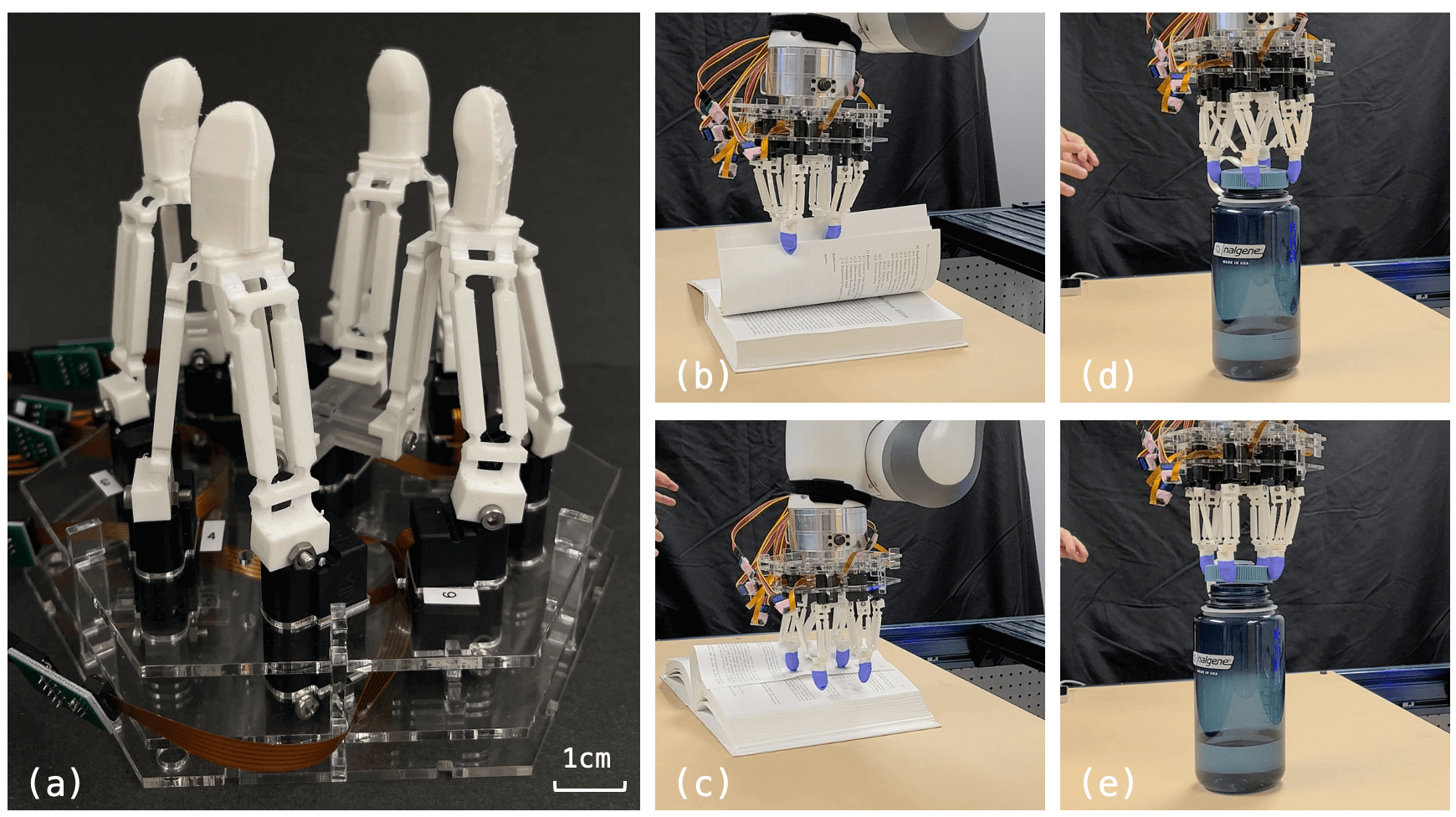}
    \caption{(a)~\namespace is a synergistic soft dexterous hand framework based on Delta robots. We demonstrate dexterous manipulation tasks such as (b)(c) turning the pages of a book and (d)(e) opening the cap of a water bottle through human teleoperation.}
    \label{fig:teaser}
    \vspace{-3mm}
\end{figure}

To achieve this goal, we need highly dexterous modular fingers.
We propose utilizing linear Delta robots, which consist of three parallel actuators connected by soft 3D-printed links and have three translational DoF with closed-form kinematics.
Multiple soft linear Delta robots are arranged in a planar layout as fingers to create dexterous hands with parallel fingertips. The parallelism and translational movements allow for easy control of the fingertips, while the soft links provide a level of compliance.
To be adaptable for different applications, we define a flexible design space that allows us to quickly and easily configure hands with different numbers of fingers and layouts, as well as different finger dimensions, workspaces, and force profiles. To reduce control complexity, we introduce hand synergies inspired by previous works~\cite{santello1998postural,ciocarlie2007dexterous,della2015dexterity} that adapted human postural synergies to robotic hands. 
Due to the inherent parallelism and modularity in linear Delta robots, we can easily configure different hand synergies by combining actuators across different fingers. 

We present~\name, a synergistic robotic hand framework for dexterous manipulation. 
By using modular Delta robots, it is easy to configure and manufacture~\namespace with low-cost components and off-the-shelf materials under \$$800$ in less than a day. In addition,~\namespace can compliantly interact with the environment due to its soft Delta links.
We provide the workspace range, kinematics accuracy, and force profile characterization data of Delta robots, which can guide the reconfiguration process for specific applications. 
We successfully demonstrated grasping daily objects using hands with two different synergies and teleoperating a hand with a robot arm for dexterous manipulation tasks. Our results show that we can easily control~\namespace by leveraging synergies while maintaining the necessary dexterity. 
Our main contributions are:
\begin{itemize}
    \item Hand design, manufacturing process, and simulation with configurable synergies in Section~\ref{sec:hand}. 
    \item Kinematics, workspace, and force profile characterization of 3D-printed soft Delta robots to assist configuring hands in Section~\ref{sec:delta}.
    \item Dexterous grasp evaluations and manipulation demonstrations using teleoperation in Section~\ref{sec:application}.
\end{itemize}

\section{Related Work}

\subsection{Robotic hands}
Robotic hands have a variety of form factors ranging from anthropomorphic hands~\cite{shadowhand, park2020open} to underactuated hands~\cite{ma2014underactuated, 9395689}, from rigid~\cite{jacobsen1984utah, allegrohand} to soft~\cite{mohammadi2020practical, abondance2020dexterous}, from tendon driven ~\cite{ma2014underactuated, fang2022multimode} to pneumatically actuated~\cite{deimel2013compliant, deimel2016novel}, and from 3D-printed~\cite{bauer2022towards, yan2022c, ma2013modular} to molded~\cite{puhlmann2022rbo}. 
However, most of these hands are difficult to reproduce or modify such as varying the number of fingers and DoF due to their complicated designs and kinematics. 
For instance, hands with tendon-driven mechanisms~\cite{catalano2012adaptive} might require an entirely new wrist design and cable routing. 
{By contrast, \namespace provides a highly flexible design space and supports a wide range of configurations including the layout, and number of fingers ($1-6$) and actuators ($3-18$) with various synergies. Different configurations of hands can be easily realized by updating the acrylic board frames and coupling bars, 3D printing the finger links as needed, and reusing all other components as shown in Fig.~\ref{fig:manufacture}. }

\subsection{Delta robots}
Delta robots~\cite{Clavel1991ConceptionDR} have been deployed for industrial pick-and-place tasks~\cite{abbdelta} and explored in research~\cite{mcclintock2018millidelta, deltaone, patil2022deltaz} because of their high speed, low inertia, and accurate kinematics. With these advantages, researchers have started using Delta robots as fingers for dexterous manipulation. A two-fingered six-DoF gripper with two Delta robots~\cite{mannam2021low} and a large array of 64 Delta robots~\cite{patil2023linear} were introduced for dexterous and distributed manipulation. Similarly, we adapt the modularized soft Delta robots as fingers for dexterous hands. We further extend it to a hand design space and propose hand synergies to reduce cost and control complexity. ~\namespace are optimized for dexterous manipulation with simple control, rather than forceful grasps that can be achieved by tendon-driven mechanisms~\cite{ma2013modular}.

\subsection{Robotic hands for dexterous manipulation}

Most soft robotic hands, especially anthropomorphic hands, are evaluated with a grasp taxonomy~\cite{7243327}. Several robotic hands are demonstrated with in-hand object repositioning~\cite{abondance2020dexterous, coulson2021elliott, puhlmann2022rbo}. We will show that~\namespace can perform versatile motions for grasping, in-hand object repositioning, and manipulating household objects. Towards learning for dexterous manipulation, ROBEL~\cite{ahn2020robel} was presented as an open-sourced platform for reinforcement learning in the real world. We provide models of \namespace both in the real world and in simulation which can be leveraged in the future for efficient skill learning. Due to dexterous hands' high dimensionalities, the control difficulty became a primary blocker for autonomous manipulation. Postural hand synergies~\cite{santello1998postural} were studied and used to reduce the dimensionality and thus simplify control~\cite{ciocarlie2007dexterous}. Researchers also adapted the synergies to hardware design~\cite{della2015dexterity, catalano2012adaptive}.~\namespace can support various software and hardware synergies based on the parallelism and modularity of linear Delta robots, and we demonstrate it with teleoperation tasks.

\section{Methodology}
\label{sec:hand} 
\namespace is a synergistic robotic hand framework to accommodate different manipulation applications. We use Delta robots as the \name' dexterous fingers which have three DoF each with simple kinematics. We provide the flexibility of dexterity by introducing actuation synergies to constrain the motions of fingers for easier control. We provide simulations of \namespace for exploring design parameters. 

\begin{figure}
    \centering
    \includegraphics[width=0.95\linewidth]{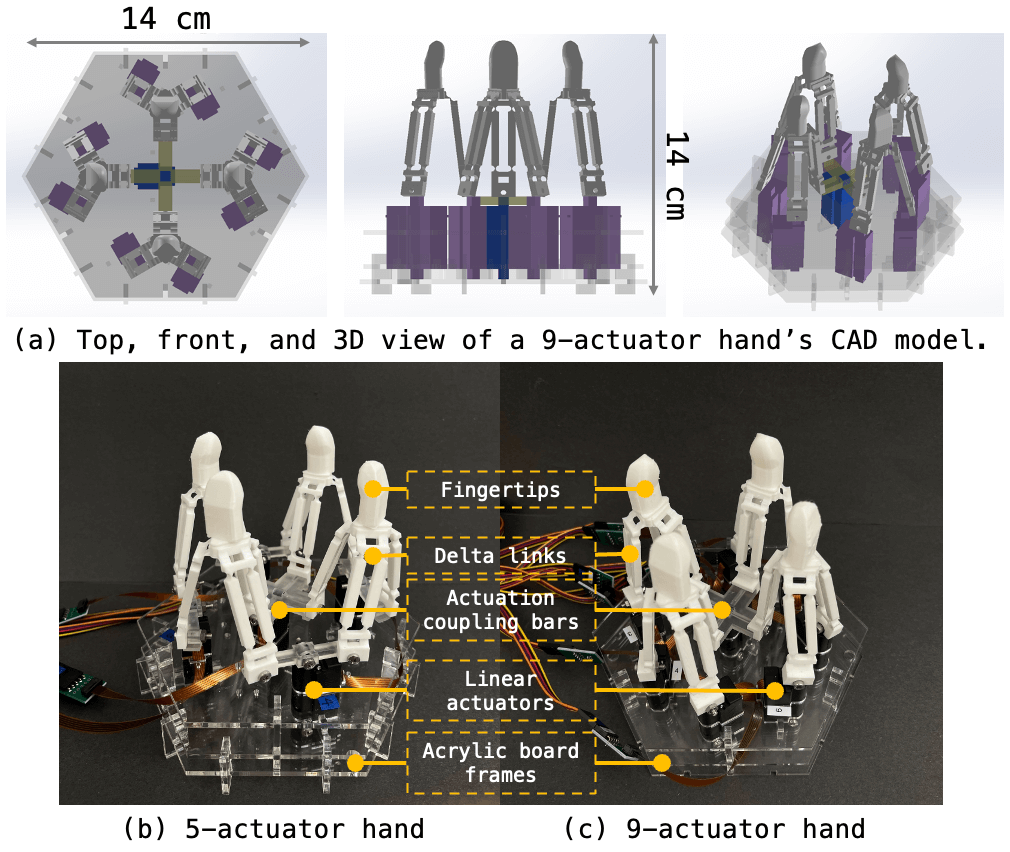}
    \caption{(a) CAD model for a 9-actuator hand where the inner actuators of four Delta robots are coupled as one center actuator (blue). Prototypes of (b) 5-actuator and (c) 9-actuator hands.}
    \label{fig:manufacture}
    \vspace{-5mm}
\end{figure}
\vspace{-2mm}
\subsection{Hand design and manufacturing}
The structure of the hand is modularized to fingertips, fingers, actuators, and frames from top to down as shown in Fig.~\ref{fig:manufacture}. This greatly simplifies hand manufacturing and re-configurations by simply modifying or replacing certain components. 
We 3D print soft fingertips with thermoplastic polyurethane (TPU) (shore hardness 95A) on an Ultimaker 3D printer, and use a pre-bent human fingertip-like shape for better small-object grasping. We adapt the compliant design of soft Delta links from~\cite{mannam2021characterization} as fingers and 3D print them with the same TPU material. We use linear actuators (Actuonix PQ12-63-6-P) with a stroke length of $20$ mm to minimize the footprint of the Delta robots for compact hand design and actuation synergies. 

As a design choice, we use four fingers arranged in a square shape to maximize the workspace of the hand. But, we can easily reconfigure the hand to one to six fingers by modifying the layout of the acrylic frames. We laser-cut three layers of acrylic boards as frames to hold the actuators in certain configurations, and
vertical fixtures to fix and reinforce the structure on the sides. Each individual actuator can be easily inserted and replaced without disassembling the hand structure. We show two hand prototypes in Fig.~\ref{fig:manufacture} (b). The size of a hand is $140 \text{ mm}\times 140\text{ mm} \times 140\text{ mm}$, which is similar to the human hand length (around 180 mm). 
{They weigh in the range of 360 g to 430 g which is smaller and lighter compared to most dexterous robotic hands~\cite{ma2013modular} given their high DoF. }
\subsection{Actuation coupling}

\begin{figure}
    \centering
    \includegraphics[width=0.99\linewidth]{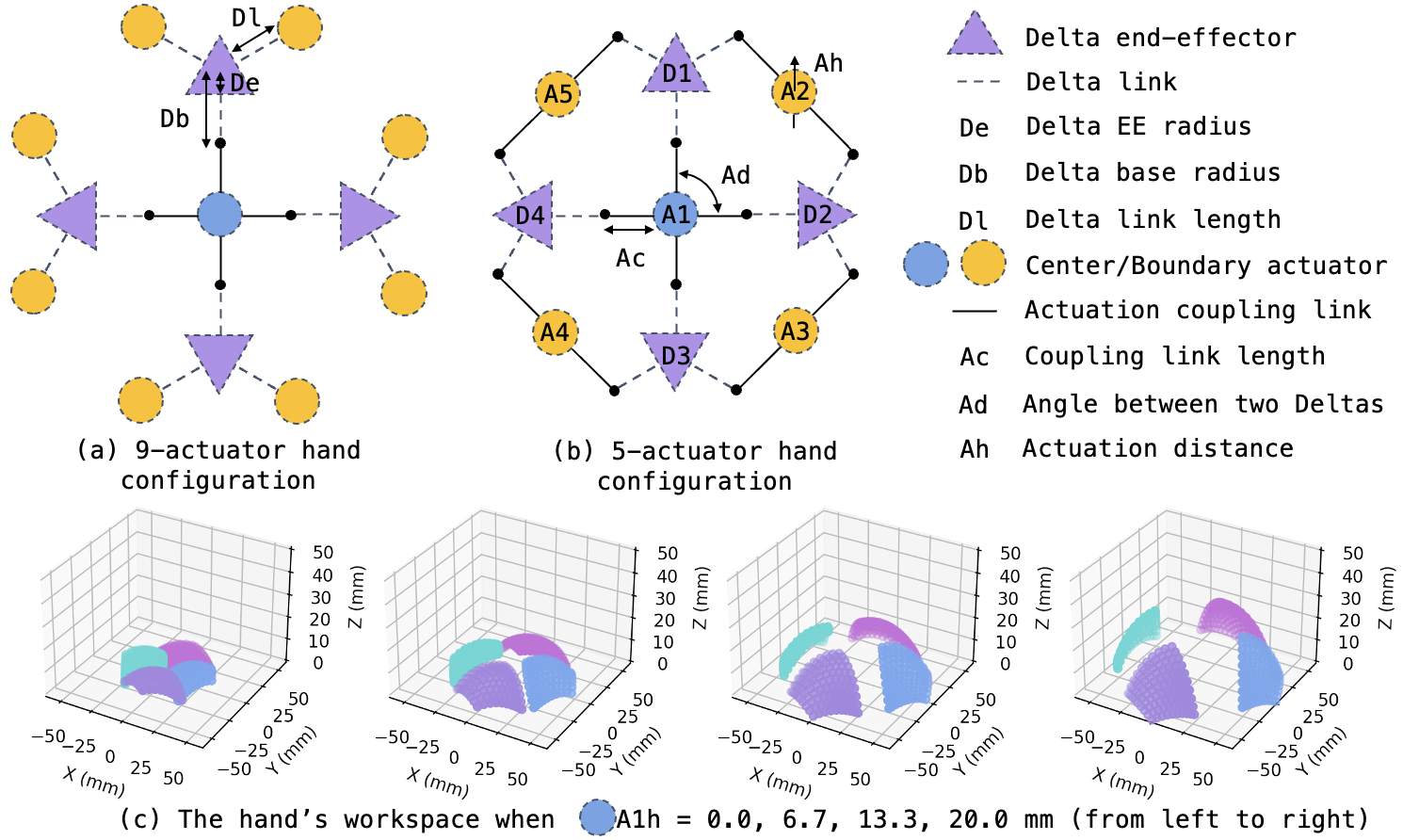}
    \caption{(a) and (b): Illustration of actuation coupling for 9-actuator and 5-actuator hands. (c) The hand's workspace varies by adjusting the center actuator's actuation distance.}
    \label{fig:coupling}
    \vspace{-5mm}
\end{figure}

A Delta robot has three DoF when it is actuated independently. However, in most manipulation tasks, motions are usually coupled through muscle synergies to create more coordinated movements between fingers~\cite{santello1998postural}. For robotic hands, along with \textit{software synergies} at the control level, \textit{hardware synergies} can be realized at the mechanical level~\cite{catalano2012adaptive}. Our hand framework can be configured to different levels of \textit{hardware synergies} by coupling the links of the Delta robots and sharing the actuators. This reduces energy consumption and control difficulty since fewer actuators are used. Here we show two potential coupling for a four-finger hand in Fig.~\ref{fig:coupling} (a) and (b). Instead of using $4 \times 3 = 12$ actuators, we can reduce the number of actuators to 9 or even 5 by coupling links across fingers while still preserving certain key DoF. Since all actuators are co-linearly arranged, the coupling can be easily achieved by modifying the frames and actuator locations. 

The study~\cite{santello1998postural} reflects that the first two principal components of postural synergies mainly control open and close motions, therefore for the 9-actuator hand, we couple the inner links from four Delta robots to one center actuator which controls the open and close motion. All other eight boundary actuators can make fingers move independently in the XY plane for finer motions. For the 5-actuator hand, besides the center coupling, each Delta robot's boundary actuators are coupled with its neighbor's. Each finger can still perform lateral motion but their neighbors will move synchronously. We show how the center actuator's motion affects the workspace of the hand in Fig.~\ref{fig:coupling} (c). When the center actuator is completely retracted, the fingers have a small overlapping workspace; the more the actuator extends, the more the fingers are separated.

\subsection{Hand parametrization}
\label{sec:parames}
~\namespace can be parameterized physically and topologically. For physical parameters, at the hand level, we define the number of fingers $N$, coupling link length $A_c$, the angle between two fingertips $A_d$, and actuation distance $A_h$. At the finger level, we define the Delta robot's end-effector radius $D_e$, base radius $D_b$, and link length $D_l$ as shown in Fig~\ref{fig:coupling}. For topological parameters, given the set of Delta robots as $\{D_1, D_2, ..., D_n\}$ and the set of actuators $\{A_1, A_2, ..., A_m\}$, we can associate them and represent actuation coupling. Such as for 5-actuator hand in Fig.~\ref{fig:coupling}, Delta robot $D_1: \{A_1, A_5, A_2\}$, actuator $A_1: \{D_3, D_4, D_1, D_2\}$.

\begin{figure}[t]
    \centering
    \includegraphics[width=0.95\linewidth]{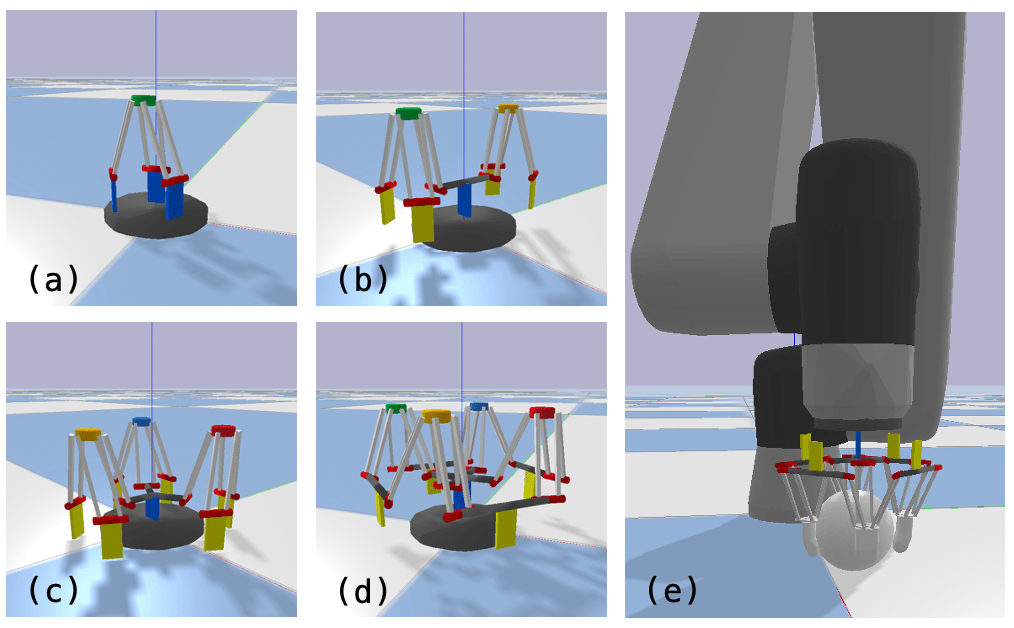}
    \caption{Simulation of~\namespace including (a) single Delta, (b) twin Deltas, (c) triple Deltas and (d) quadruple Deltas as robotic hands. (e) Integration of a four-finger hand on a robot arm.}
    \label{fig:simulation}
    \vspace{-5mm}
\end{figure}
\subsection{Hand kinematics}
Single Delta robot's kinematics have closed-form unique solutions that we can adapt to Delta hands. For each Delta robot $D_i$, we define its actuation space as $A_{D_i} = \{a_{i1}, a_{i2}, a_{i3}\} \in \mathbb{R^{+}}$, and end-effector space as $E_{D_i} = \{x_{i1}, y_{i2}, z_{i3}\}\in \mathbb{R^{+}}$, then we have forward and inverse kinematic models as $E_{D_i} = FK(A_{D_i})$, and $A_{D_i} = IK(E_{D_i})$. For a Delta hand without actuation coupling, we can simply use a single Delta robot's kinematics model for each finger. For hands with actuation coupling, all the \textit{hardware synergies} are \textit{rigid synergies}~\cite{catalano2012adaptive}. Thus we can control and model all DoF with kinematics models. Forward kinematics are the same by setting each Delta robot's actuation with its corresponding actuators' values. However, for the inverse kinematics, since the end-effector space is constrained by the reduced actuation dimensions, there might not be a precise solution for the desired end-effector position. Thus, we still first solve individual Delta robot's IK given the desired end-effector position, then project the solution to the feasible actuation space $\widetilde{A} = P A$. 
Such as for the 9-actuator hand, the projection matrix P can be defined as:
\begin{equation}
P = \begin{bmatrix}
\frac{1}{4} & \frac{1}{4} & \frac{1}{4} & \frac{1}{4} & 0_{1 \times 8}\\
0_{8 \times 1} & 0_{8 \times 1} & 0_{8 \times 1} & 0_{8 \times 1} & I_{8 \times 8} \\
\end{bmatrix} \in \mathbb{R}^{9 \times 12}
\label{eq:projection}
\end{equation}
We approximate the hand's IK solution by averaging the center actuation distance. The projection matrix reduces the actuation space dimension to reduce the control complexity which is similar to Eigengrasps~\cite{ciocarlie2007dexterous}. Although the solution might not be precise, we demonstrate in Section~\ref{sec:application} that we can still successfully complete dexterous grasping and manipulation tasks with the help of compliance in the finger links.
\begin{figure*}[ht]
    \centering
    \includegraphics[width=0.85\linewidth]{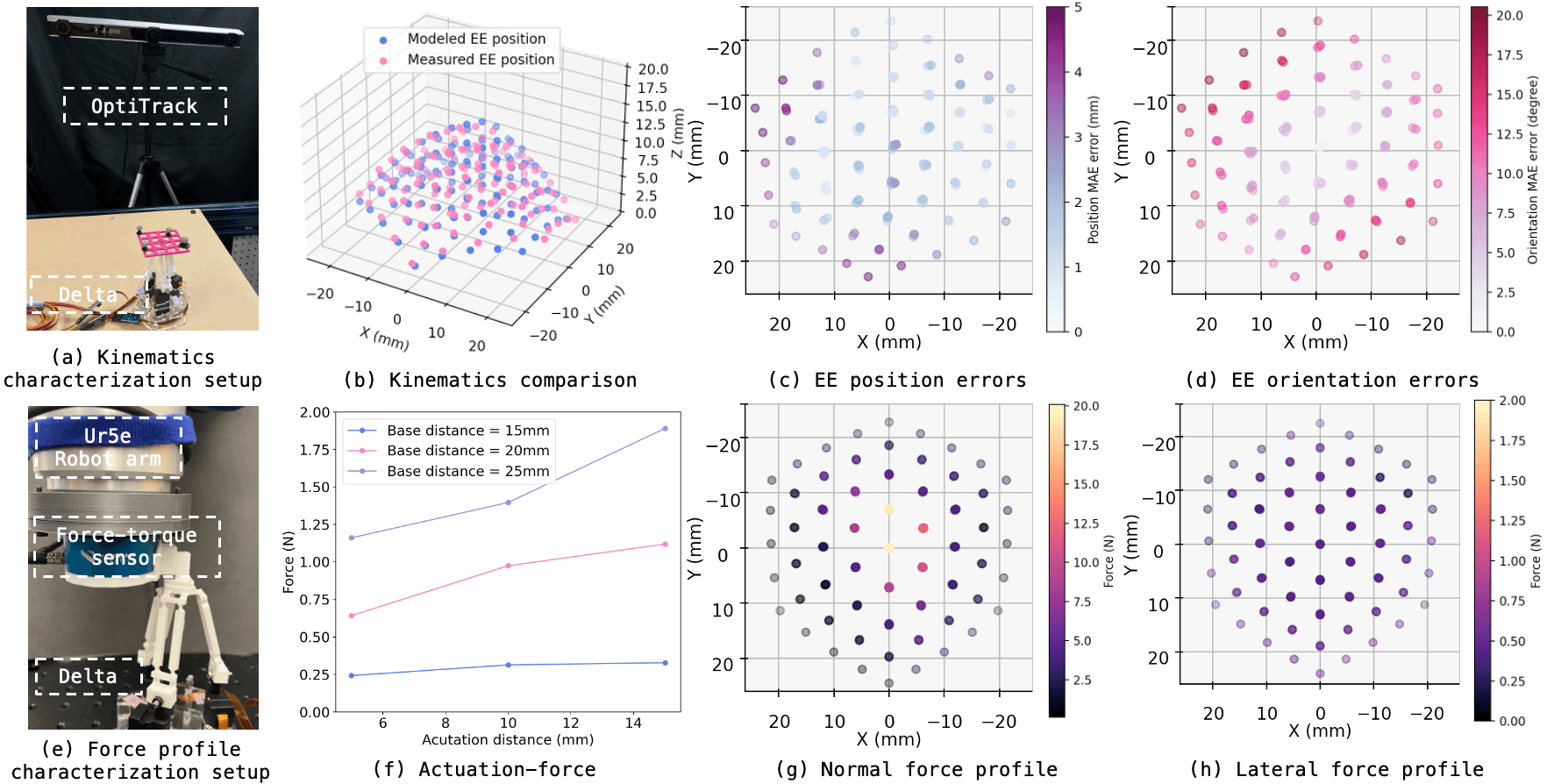}
    \caption{(a) Kinematics characterization setup with an OptiTrack system to track a Delta robot's end-effector (EE) pose. (b) Comparing the modeled EE's positions from forward kinematics and recorded positions from the OptiTrack system over a Delta robot's workspace. (c) The EE's position errors and (d) orientation errors. (e) Force profile characterization setup with a UR5e robot arm to execute Cartesian motions, a force-torque sensor to record contact forces, and a Delta robot fixed on the table to actuate towards the contact direction. (f) Correlation between actuation distances and contact forces. (g) (h) Characterization of a Delta robot's force profile in normal and lateral directions. }
    \label{fig:kinematics}
    \vspace{-6mm}
\end{figure*}
\subsection{Hand design in simulation}

Although we just presented two hand designs in the aforementioned section, we can build various hands based on different synergies. A practicable simulation enables us to explore unlimited design parameters, and iterate and validate the design in a fast and safe manner. We use PyBullet~\cite{coumans2016pybullet} and provide a high-level urdf generator given the hand parameters defined in Section~\ref{sec:parames}. As shown in Fig.~\ref{fig:simulation}, we can build hands with anywhere from one to six fingers and simulate different hand synergies. We also integrate \namespace with a robot arm for grasping evaluations and can use it for potential design optimization and policy learning in simulation.

\section{Delta Robot Characterization}
\label{sec:delta}
Serving as the functional "fingers" of the robotic hand, it is crucial to characterize and optimize the Delta robots' capabilities. 
Our objective is to maximize the force profile and workspace of the fingers based on functionality, and simultaneously, we strive to minimize the footprint and weight of the hand for efficient mounting on a robot arm, taking feasibility into account. Previous work~\cite{mannam2021characterization} has characterized the soft materials of Delta robots but they didn't explore optimization of the structure. In the following sections, we characterize the kinematics, workspace, and force profile of Delta robots and show how the design parameters affect the behaviors of Delta robots. This will assist with hand configurations for different applications.
\subsection{Kinematics}

We compare the kinematics accuracy of a Delta robot from the modeled and the observed end-effector's poses. For both modeling and recording, we discretize the actuation space $\mathbb{R}^3$ for the three linear actuators into $5\times 5\times 5$ grids with $4$ mm granularity given our $20$ mm stroke actuators. We solve the forward kinematics and get modeled end-effector positions. Since Delta robots have only three translational DoF, the modeled orientations are all aligned with the principal axes. For observing poses, we use an OptiTrack \cite{OptiTrack} system to track the end-effector poses when actuating motors to the pre-defined distances (Fig.~\ref{fig:kinematics} (a)). Then we compare each pair of predicted vs recorded poses (Fig.~\ref{fig:kinematics} (b)). The average translational mean absolute errors (MAE) over the workspace are $0.73$ mm, $0.77$ mm, and $0.43$ mm, and the average orientation MAE errors are $3.36^{\circ}$, $2.28^{\circ}$, and $3.97^{\circ}$ along the X, Y, and Z axes respectively. {The kinematics precision is higher than that reported for the Stewart Hand's~\cite{9395689}.} As shown in Fig.~\ref{fig:kinematics} (c) and (d), both translational and orientational accuracy decrease towards the workspace's boundary where the soft Delta robot undergoes larger internal forces leading to more deformation as compared to the center of the workspace.

\subsection{Force profile}

We characterize a Delta robot's force profile in four pre-defined lateral direcitons (+X, -X, +Y, -Y) and one normal direction (-Z) using a pre-defined actuation distance $5$ mm, and the correlations between the actuation distances and forces under three different configurations.
As shown in Fig.~\ref{fig:kinematics} (e), the force profile characterization setup includes a Delta robot fixed on a tabletop optical board and a 6-Axis force-torque sensor mounted on a UR5e robot arm. We customize the end-effector (EE) of the Delta robot to be a cube and a flat surface on the force-torque sensor to constrain the contact face. We define the same $5\times 5\times 5$ grids in actuation space as the kinematics characterization. For each grid, the Delta robot first actuates to the desired location. The UR5e robot arm approaches the Delta robot from a safe distance and stops when a light contact ($0.2$ N) is detected by the force-torque sensor. Then the Delta robot actuates towards the touch direction $5$ mm more and holds for $5$ seconds. The force readings are recorded from the force-torque sensor during the experiments. 

\subsubsection{Lateral and normal force profile over workspace} We plot the force profiles given $5$ mm actuation in Fig.~\ref{fig:kinematics} (g) and (h). The peak of normal forces is at the center of the workspace and reaches up to $20$ N. The forces decrease towards the boundary of the workspace. The average normal force is $5.32$ N. In the lateral directions, similarly, the Delta robot becomes less stable towards the boundary, and the average of lateral forces are $0.61$ N, $0.59$ N, $0.76$ N, and $0.67$ N along +X, -X, +Y, -Y axes. This is because the parallel links become less supportive when the Delta robot's EE moves toward the boundary of the workspace. {We then measure the maximum lateral forces of a Delta robot to indicate the grasping forces of \name. We choose four points along the finger closing direction within the hand and actuate the Delta robot until it buckles. The maximum forces from the outer to inner locations are $1.49$ N, $2.34$ N, $2.21$ N, and $1.92$ N. These values are on the same force level as the soft hand of Abondance et al.~\cite{abondance2020dexterous}.}

\subsubsection{Force profile given different actuation distance} The actuation distance also affects the contact force since Delta robots are soft and compliant. We test the lateral direction forces by actuating the Delta robot $5$ mm, $10$ mm, and $15$ mm starting from the home position. We also test three different configurations by adjusting the base radius ($D_b$) of the Delta robot to $15$ mm, $20$ mm, and $25$ mm. As shown in Fig.~\ref{fig:kinematics} (f), the averaged lateral direction forces are linearly correlated to the actuation distances and the coefficient factor varies with different base distances.

\subsection{Durability}
To evaluate the lifespan, we actuate and reset a Delta robot 10,000 times (with the same $5\times 5\times 5$ grids in the workspace for 80 times) and then re-test the kinematics and force profile. We report $0.90\text{ mm, } 0.76\text{ mm, } 0.38\text{ mm,}$ as average translational MAE errors, $3.64^{\circ}$, $2.55^{\circ}$, $3.80^{\circ}$ as average orientation MAE errors, and $0.73$ N as the average lateral force profile. These are on the same level as the previous characterization performance indicating no degradation of the Delta robot.

\subsection{Workspace}

\begin{figure}[h]
    \centering
    \includegraphics[width=1.0\linewidth]{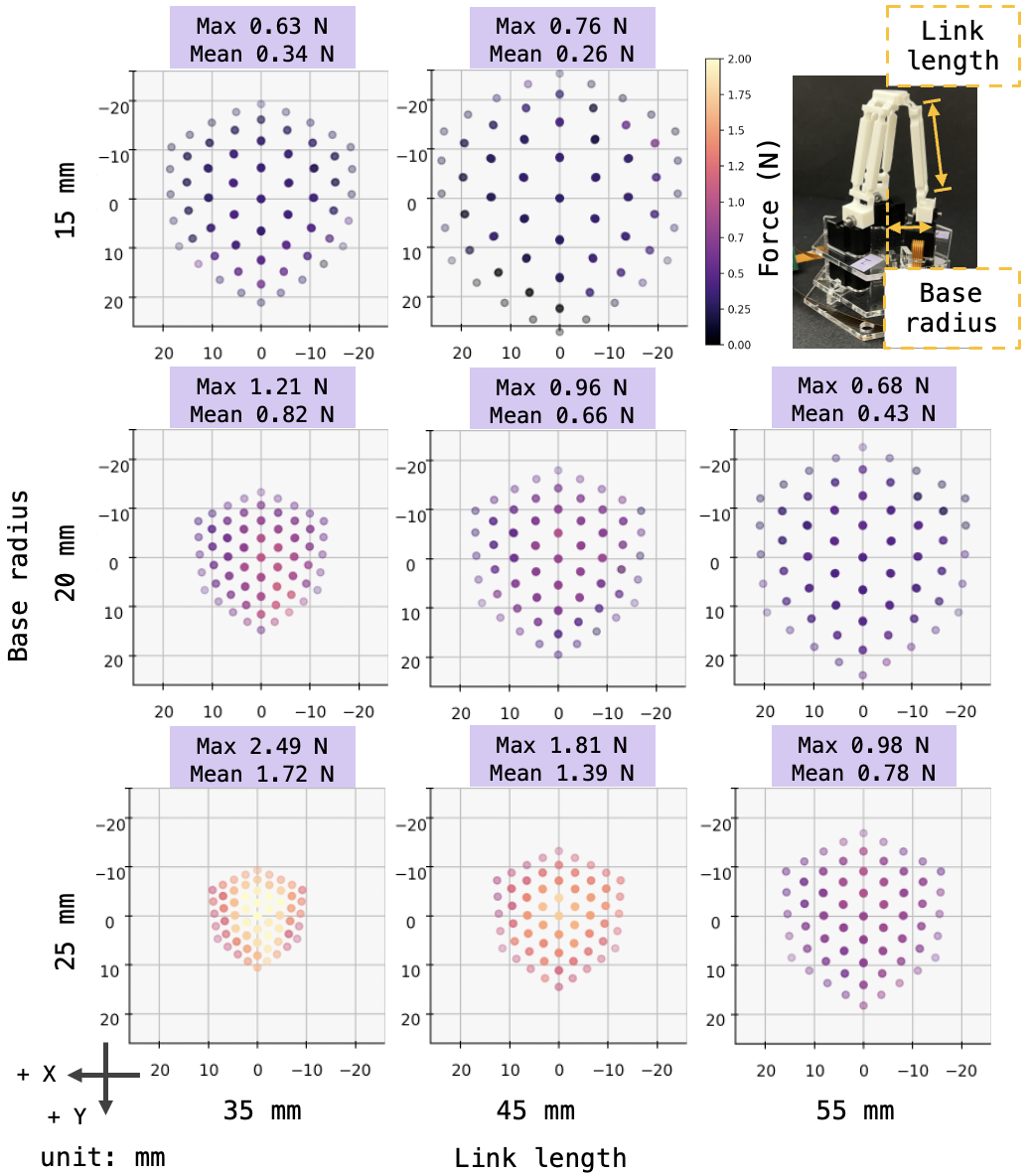}
    \caption{Characterization of force profile and workspace size by varying the Delta robot's base radius and link length. With a smaller base radius and longer link length, the workspace increases but the force profile decreases, and vice versa.}
    \label{fig:force-all}
    \vspace{-6mm}
\end{figure}

To minimize the hand size while keeping a reasonable workspace and force profile for most manipulation tasks, design parameters including Delta robot's link length ($D_l$), base radius ($D_b$), and EE radius ($D_e$) as indicated in Fig.~\ref{fig:coupling} can be varied and configured. Here we characterize the workspace size and force profile by configuring the base radius as $15$ mm, $20$ mm, and $25$ mm, and link length as $35$ mm, $45$ mm, and $55$ mm. We keep the same EE radius as 6 mm to fit the fingertip. We follow the same procedure as force characterization. We plot the modeled workspace and recorded averaged lateral direction force profile data in Fig.~\ref{fig:force-all}. We observe the trade-off between the workspace size and stability of the Delta robot.

For household object manipulation in Section~\ref{sec:application}, we choose to use Delta robots with a $20$ mm base radius ($D_b$) and a $45$ mm link length ($D_l$) to accommodate the size and weight of the objects. We also pre-define the coupling link length ($A_c$) as $20$ mm and the angle between fingers (${A_d}$) as $90 ^{\circ}$ to let the fingertips overlap when they are fully closed for firm small object grasping as shown in Fig.~\ref{fig:coupling} (c). This gives us a maximum hand workspace of $124\text{ mm}\times 124\text{ mm}\times 25\text{ mm}$.

\section{Dexterous grasping and manipulation}
\label{sec:application}
\subsection{Evaluation of grasping}
We evaluate~\name' grasp capabilities with various objects qualitatively in the real world and quantitatively in simulation for two different hand configurations.

\subsubsection{Grasp demonstration}
\begin{figure*}[t]
    \centering
    \includegraphics[width=0.95\linewidth]{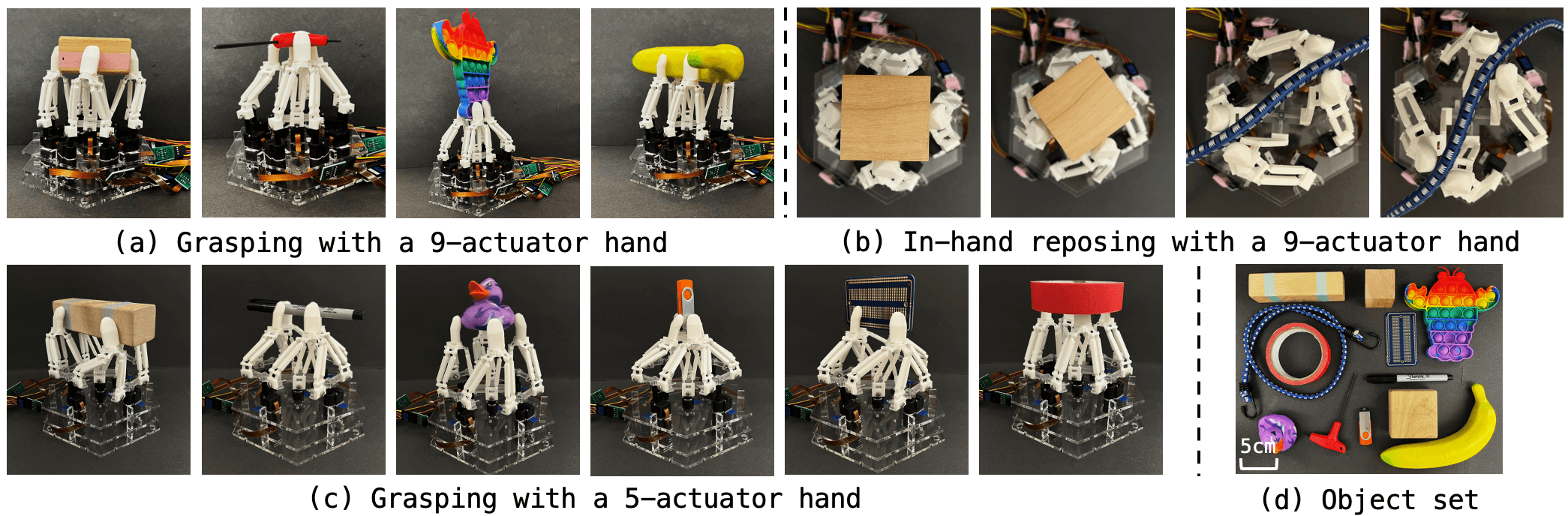}
    \vspace{-1mm}
    \caption{Grasping gallery with~\name. We grasp (d) twelve daily objects with (a) a 9-actuator hand and (c) a 5-actuator hand. We also show (b) in-hand cube and rope repositioning.}
    \label{fig:grasp}
    \vspace{-3mm}
\end{figure*}

\begin{table*}[t]
\centering
\setlength{\tabcolsep}{4pt}
\renewcommand{\arraystretch}{1.2}
\begin{tabular}{llccccccccccccc}
\Xhline{2\arrayrulewidth}
{Object} & {} & {\includegraphics[height=.05\textwidth]{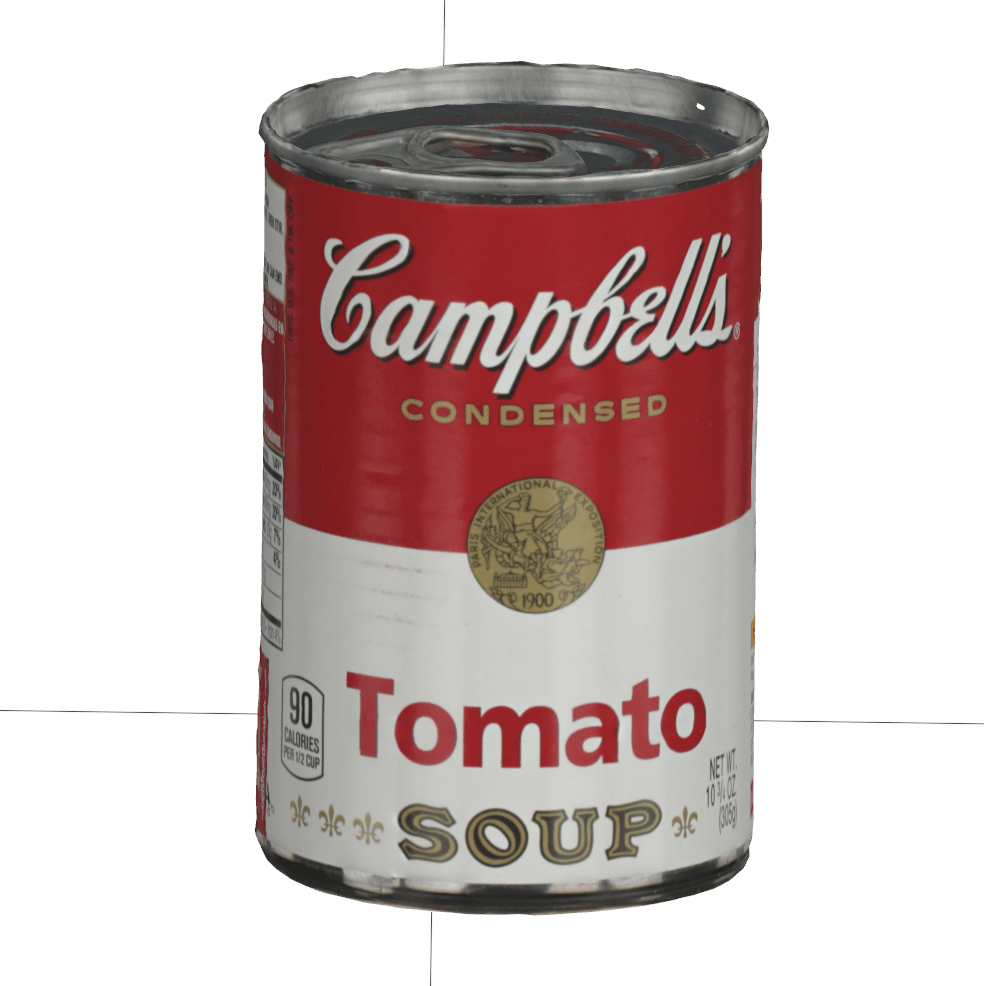}} & 
{\includegraphics[height=.05\textwidth]{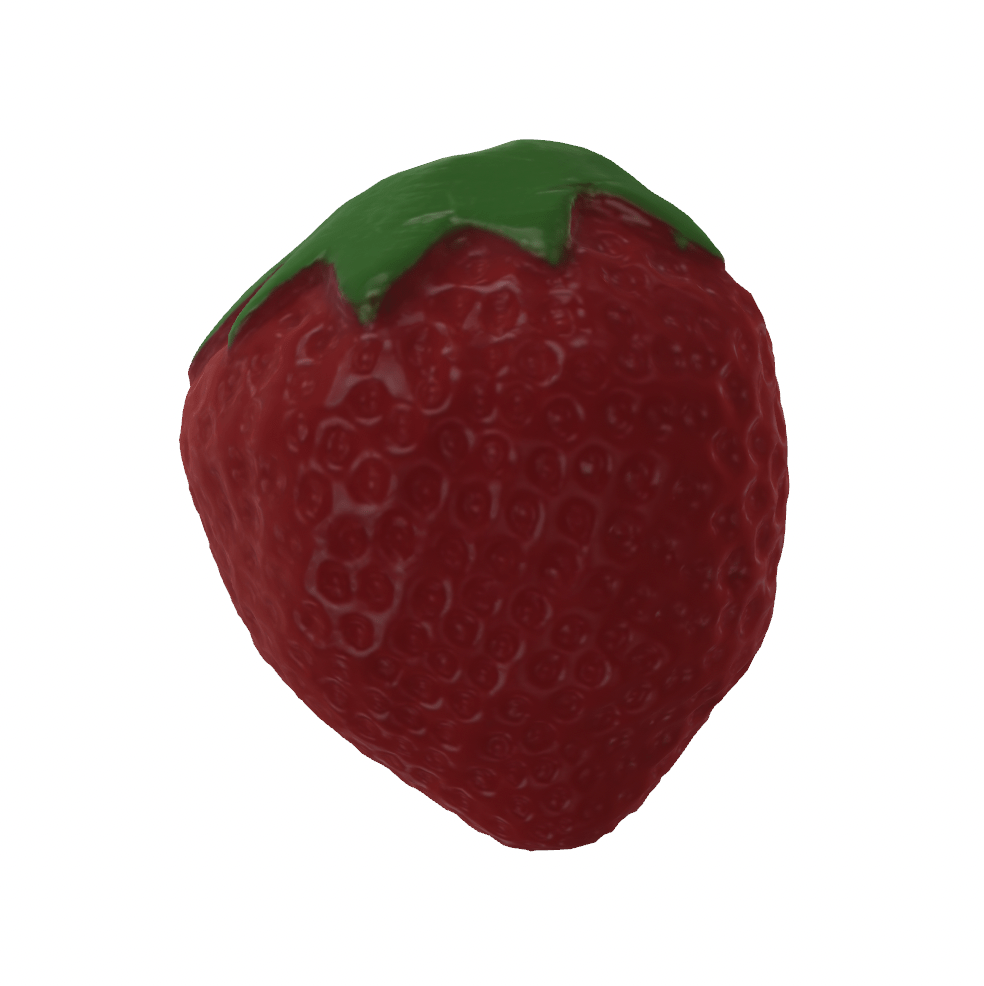}} &
{\includegraphics[height=.05\textwidth]{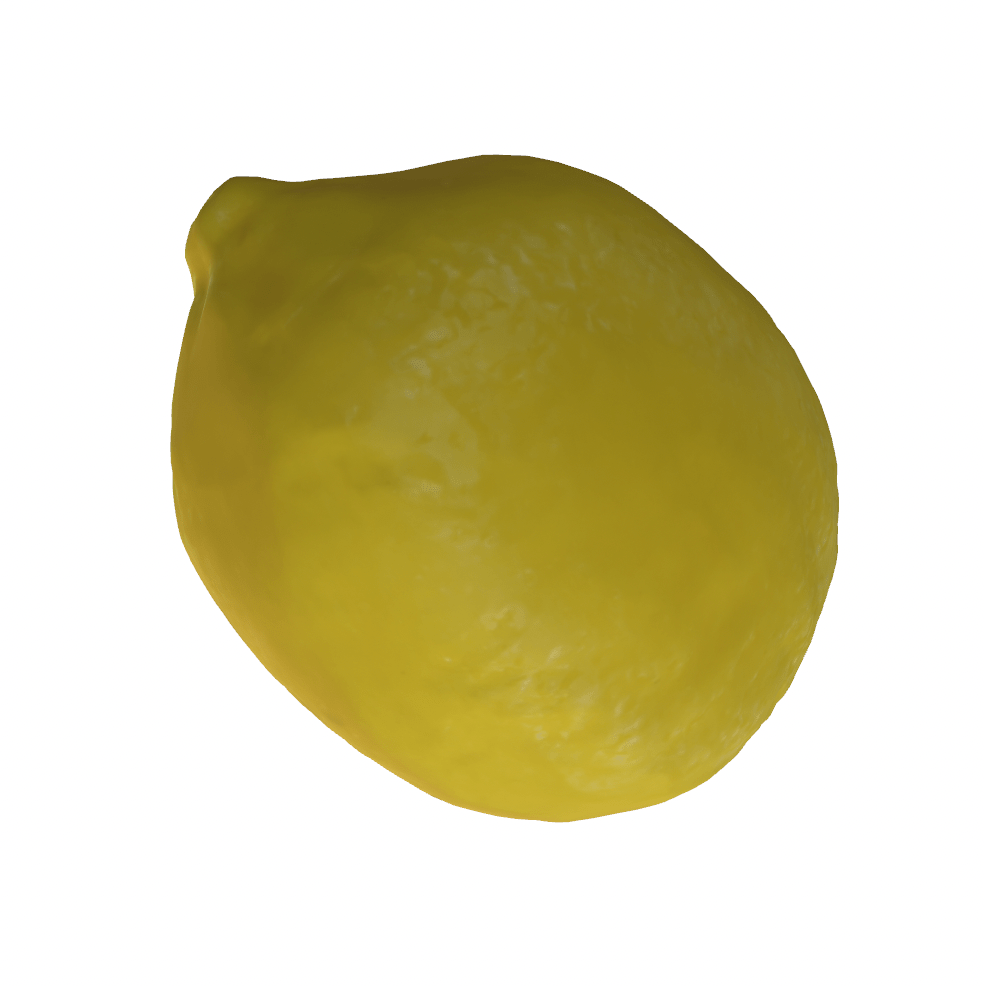}} & 
{\includegraphics[height=.05\textwidth]{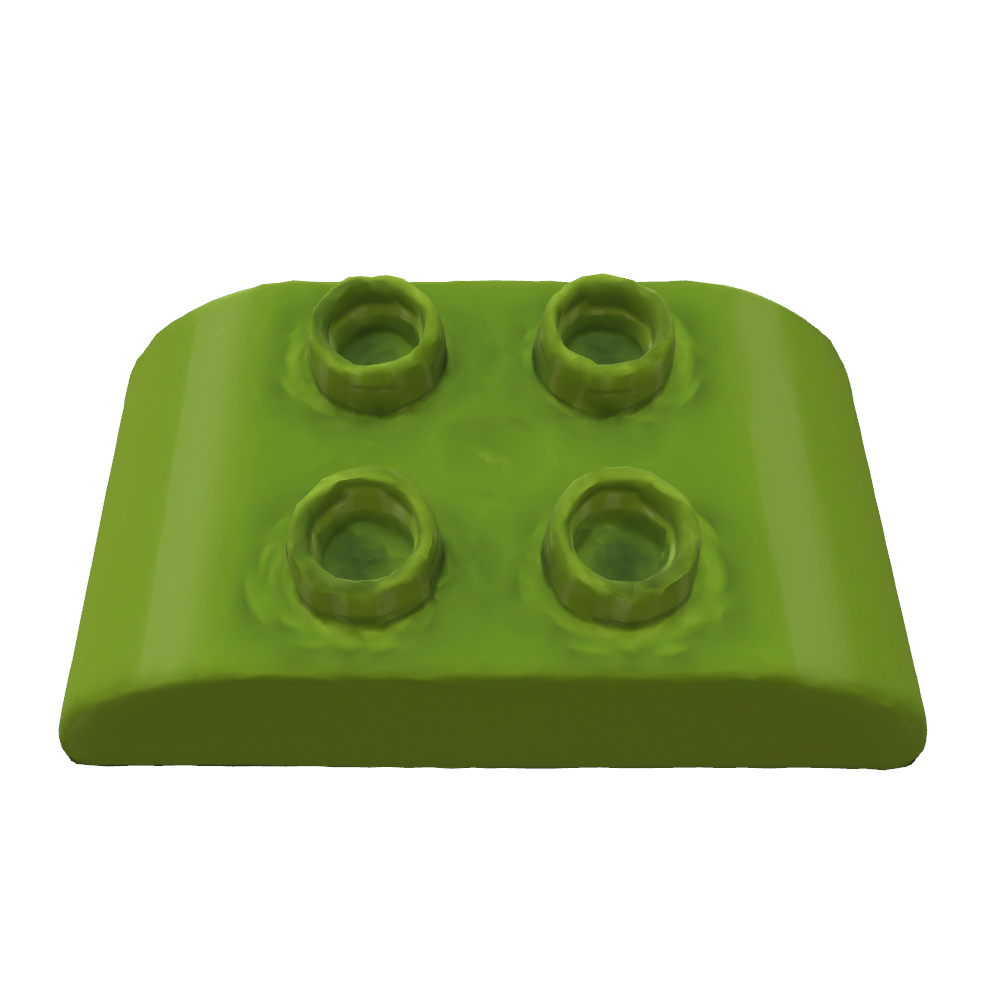}} &
{\includegraphics[height=.05\textwidth]{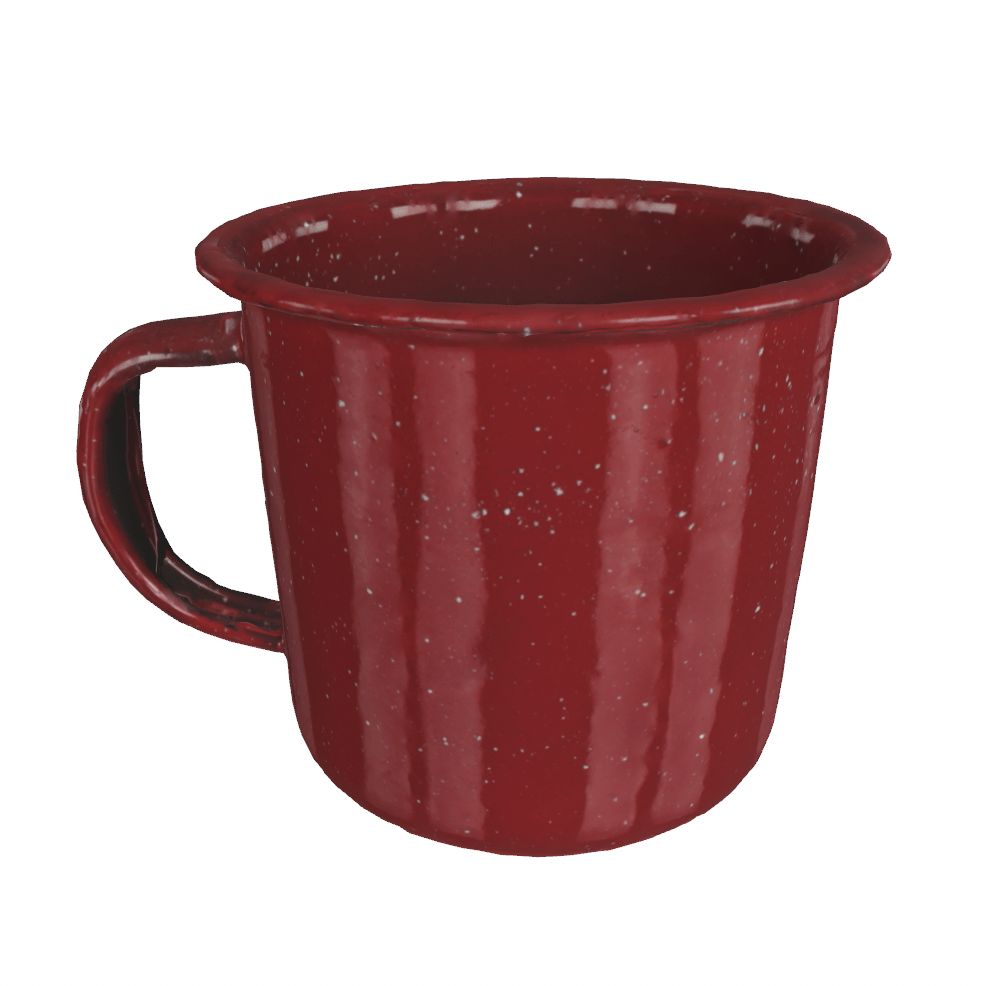}} & 
{\includegraphics[height=.05\textwidth]{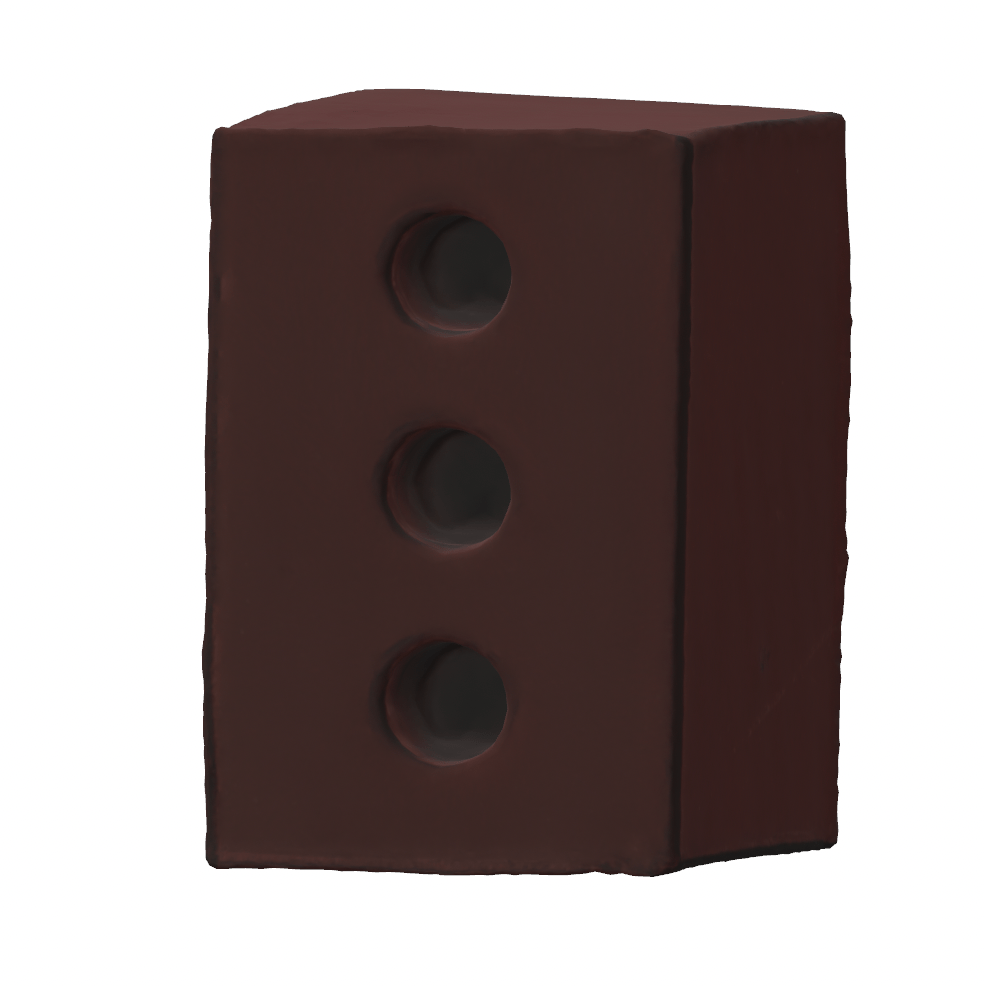}} &
{\includegraphics[height=.05\textwidth]{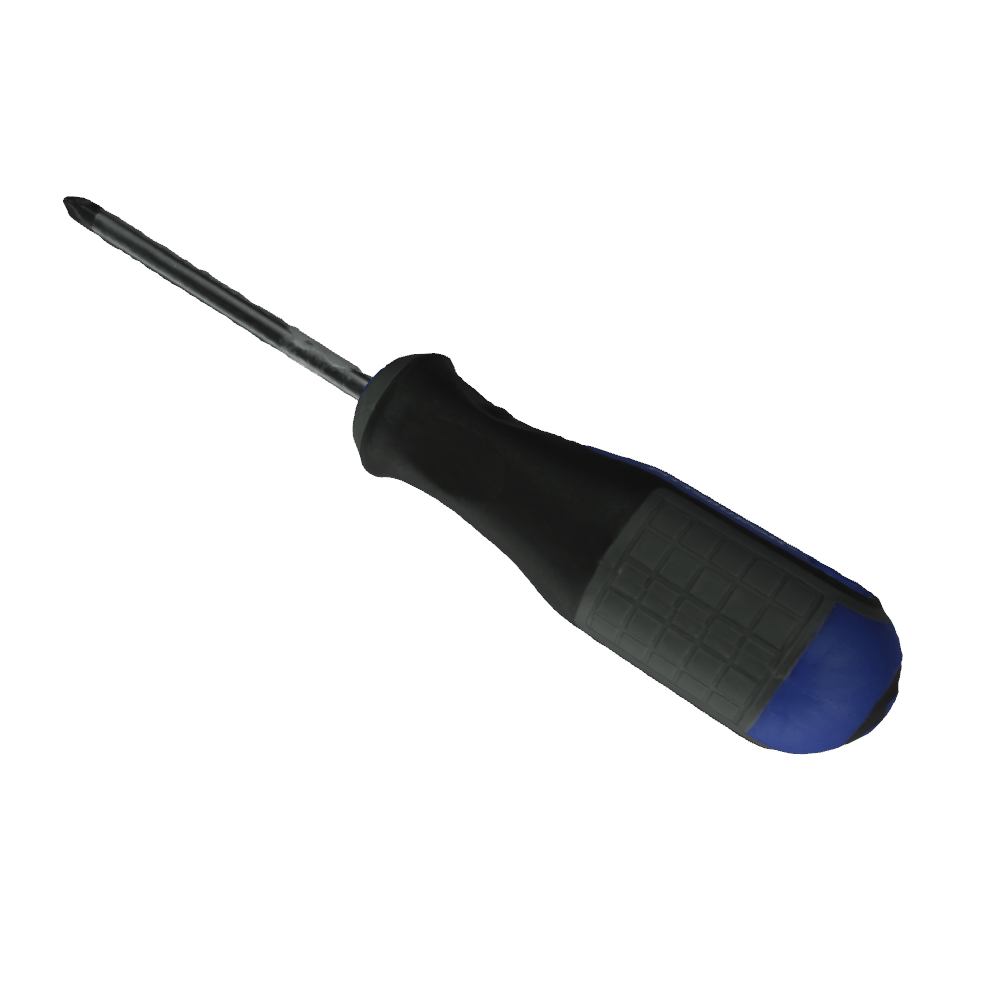}} &
{\includegraphics[height=.05\textwidth]{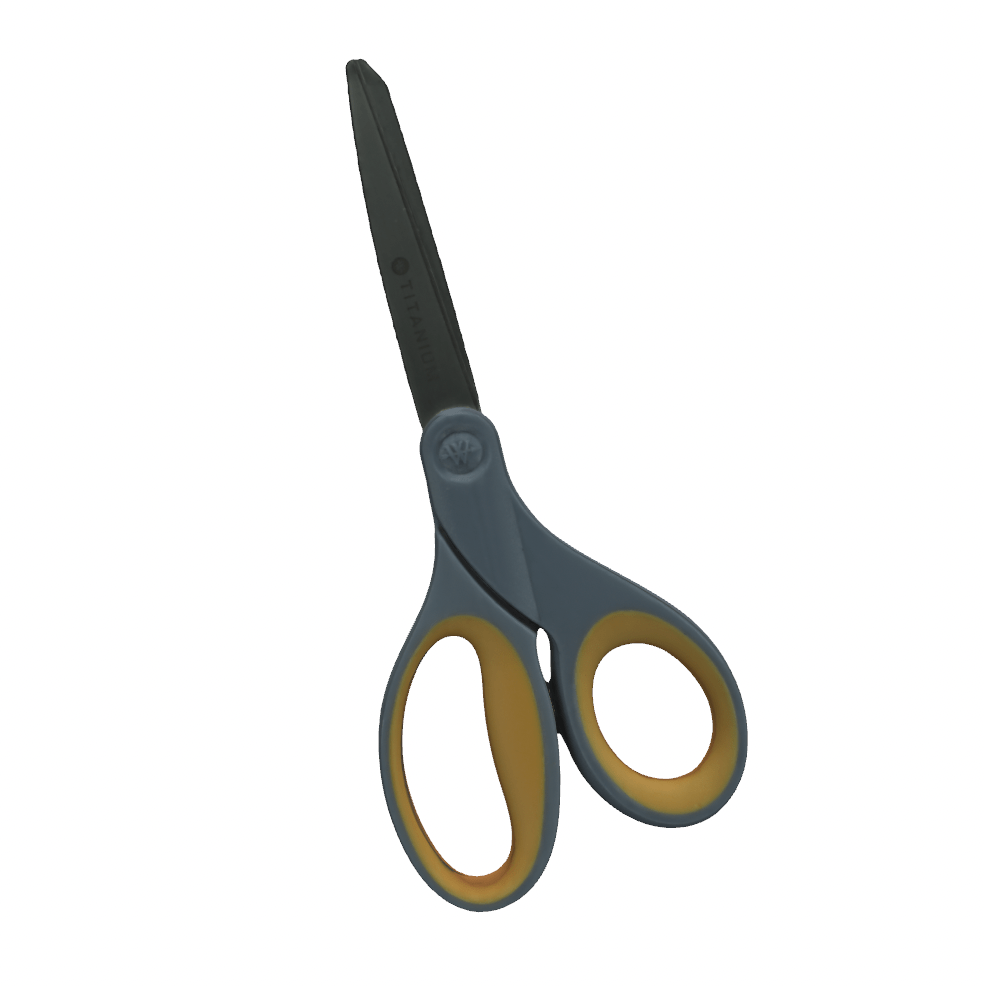}} &
{\includegraphics[height=.05\textwidth]{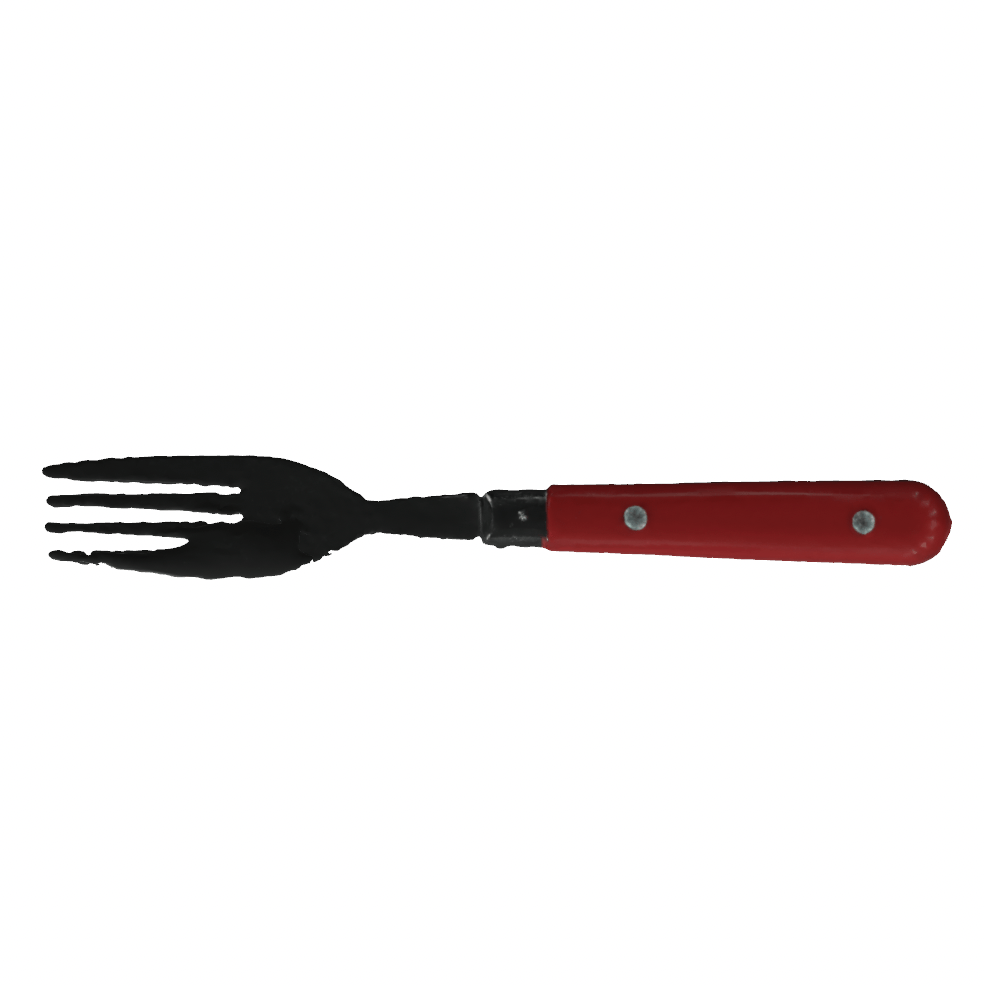}} & 
{\includegraphics[height=.05\textwidth]{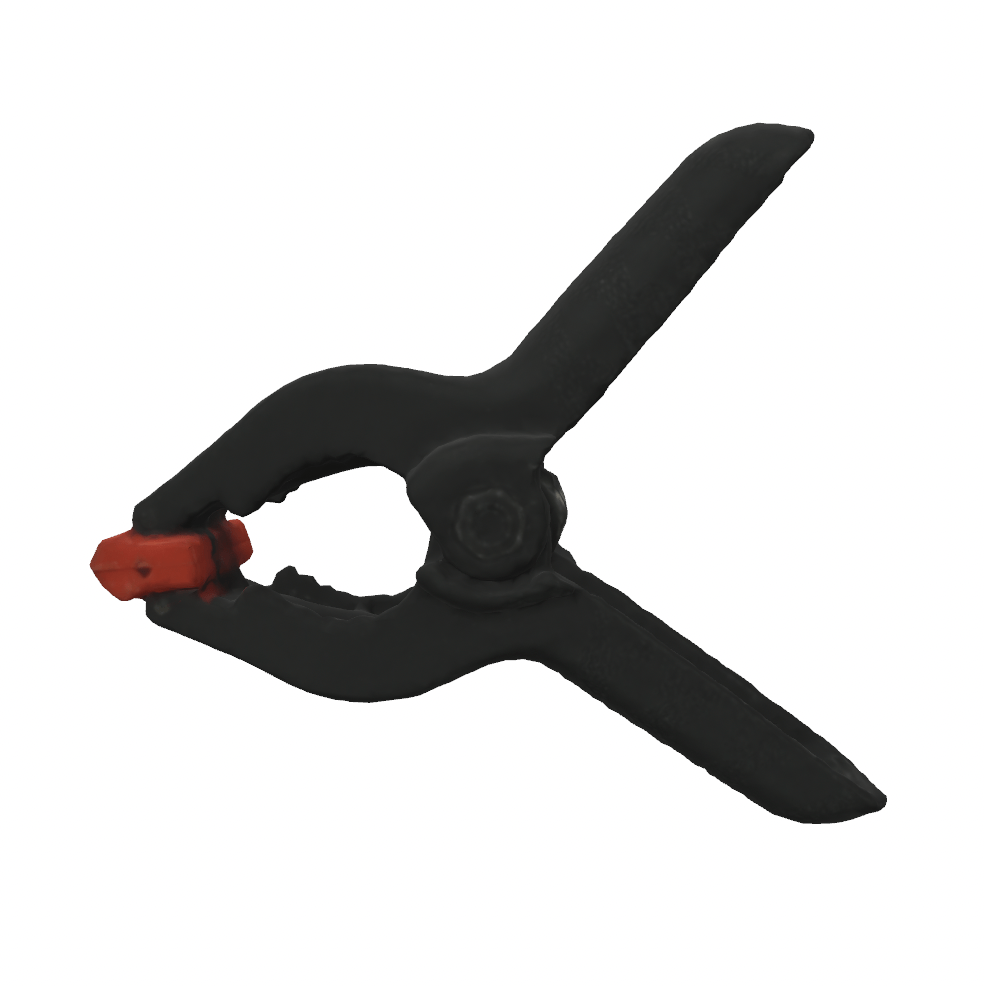}} &
{\includegraphics[height=.05\textwidth]{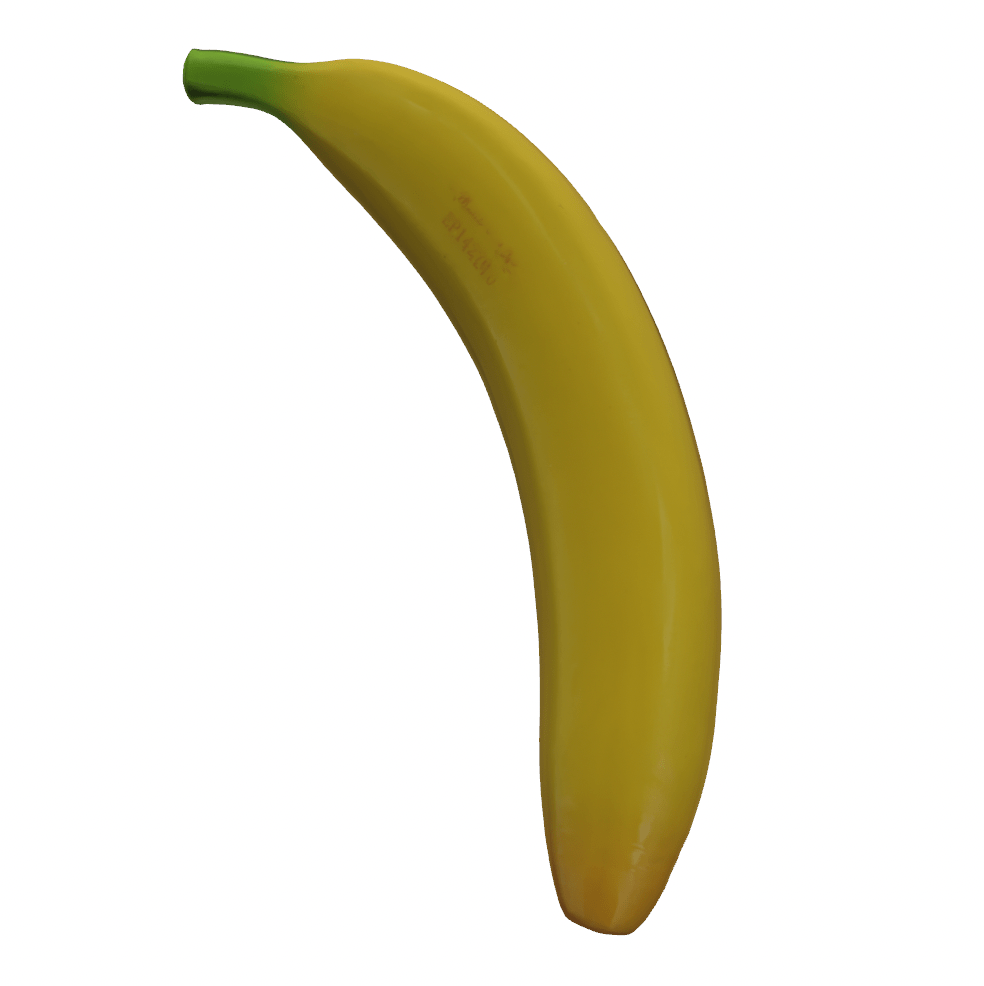}} &
{\includegraphics[height=.05\textwidth]{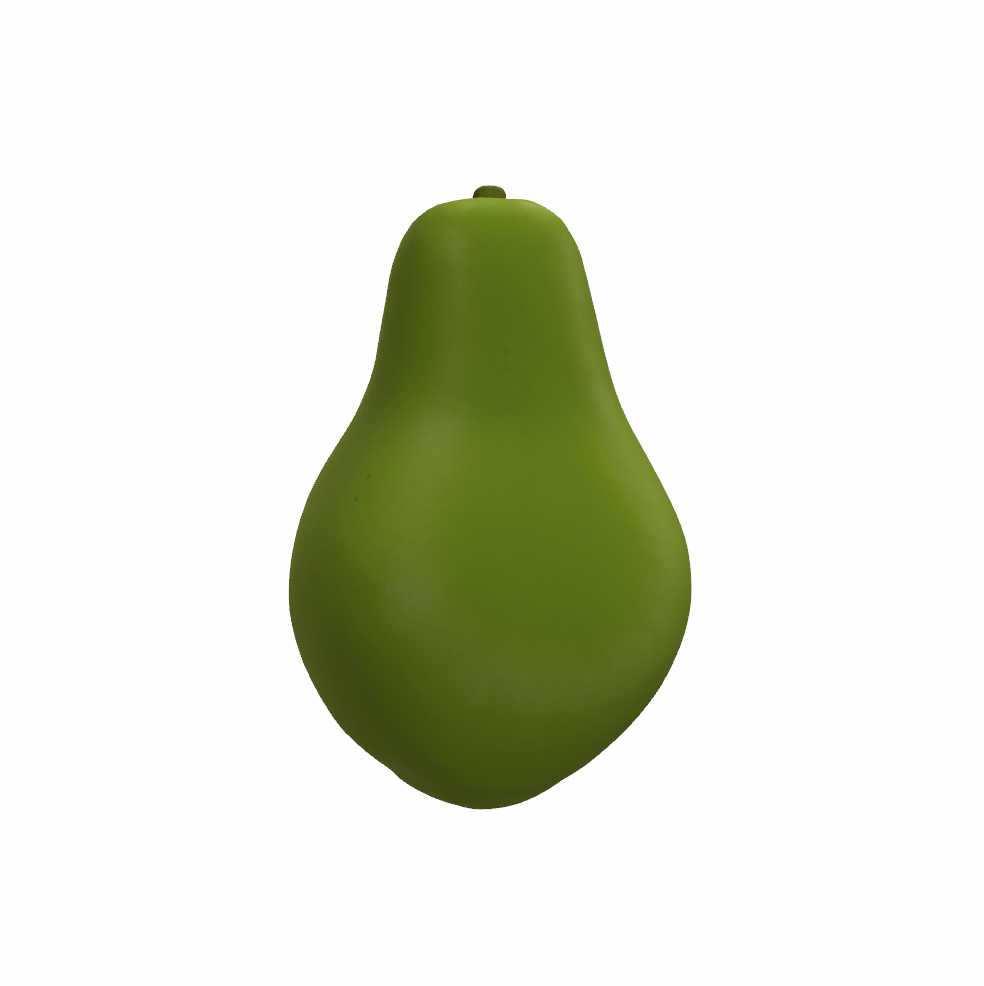}} 
 \\ \hline
{\# force closure} & 5-actuator & 720 & 1110 & 760 & 1025  & 291  & 1098 & 1034 & 564 & 765 & 809 & 804 & 412 \\ 
{} & 9-actuator  & 586 & 872 & 643 & 881  & 218  & 926 & 942 & 516 & 661  & 708 & 674 & 386 \\
\hline
{Grasping score} & 5-actuator &  130.216   & 95.642  & 102.964 &  92.776 & 131.036 & 126.796 & 63.981 & 76.483 & 76.442  & 85.983 & 75.161 & 23.067\\ 
{($Q_{LRW}$)} & 9-actuator &  132.105   & 95.312 & 115.514 & 95.616  & 128.247 & 122.690 & 64.350 & 65.742 & 74.753 & 88.252 & 52.418 & 48.967\\
\Xhline{2\arrayrulewidth}
\end{tabular}
\caption{Grasping evaluation of 5- and 9-actuator hands on 12 objects from the YCB dataset. We use the number of force closure grasps found out of 20,000 random searches and the largest-minimum resisted wrench ($Q_{LRW}$) grasping metric to evaluate the hand's grasping capability. We observe that the 5-actuator hand has better sampling efficiency with lower actuation space dimension while having similar-quality grasping compared with the 9-actuator hand.}
\label{table:ycb-evaluate}
\vspace{-5mm}
\end{table*}

Here we show grasping a set of 12 objects with 9-actuator and 5-actuator hands in the  Fig.~\ref{fig:grasp} (a)(c). We select objects (Fig.~\ref{fig:grasp} (d)) with various sizes, weights, and geometries to qualitatively show the diversity of grasps. Along with static grasping, we can also reposition objects in hand (a cube and a rope) (Fig.~\ref{fig:grasp} (b)). 

\subsubsection{Evaluation on YCB object dataset with grasp metrics}

To compare 5-actuator and 9-actuator hands' grasp capability, we evaluate them on the YCB object dataset~\cite{calli2017yale} with grasp metrics~\cite{roa2015grasp} in simulation. First, we generate a pre-grasp shape distribution by only adjusting the open and close motions of the hand and the pose of the hand to get in contact with the object. Then, for different actuator coupling configurations, we fine-tune the grasp shape by sampling each actuator's actuation individually based on the pre-grasp shape distribution. For each sampled grasp configuration, we evaluate whether the grasp is a force closure by considering the contact friction with a friction coefficient of $0.5$. Then for force-closure grasps, we calculate the grasp quality scores by using the largest-minimum resisted wrench ($Q_{LRW}$) metric~\cite{roa2015grasp}.

We choose 12 objects with various shapes and sizes for evaluation. We position the hand at 20 different heights and 4 different orientations ($0^{\circ}$, $45^{\circ}$, $90^{\circ}$, $135^{\circ}$ around the z-axis), and average the scores as the final evaluation scores. We show the number of force closures found from 20,000 random samples, and the grasping scores for both hands in Table.~\ref{table:ycb-evaluate}. 
We observe more force closures found for the 5-actuator hand on all objects indicating better sampling efficiency, and similar-level grasping scores for both hands which means that even with fewer DoF, it is still possible to grasp these objects.

\subsection{Teleoperation for dexterous manipulation}

\begin{figure*}[t]
    \centering
    \includegraphics[width=0.95\linewidth]{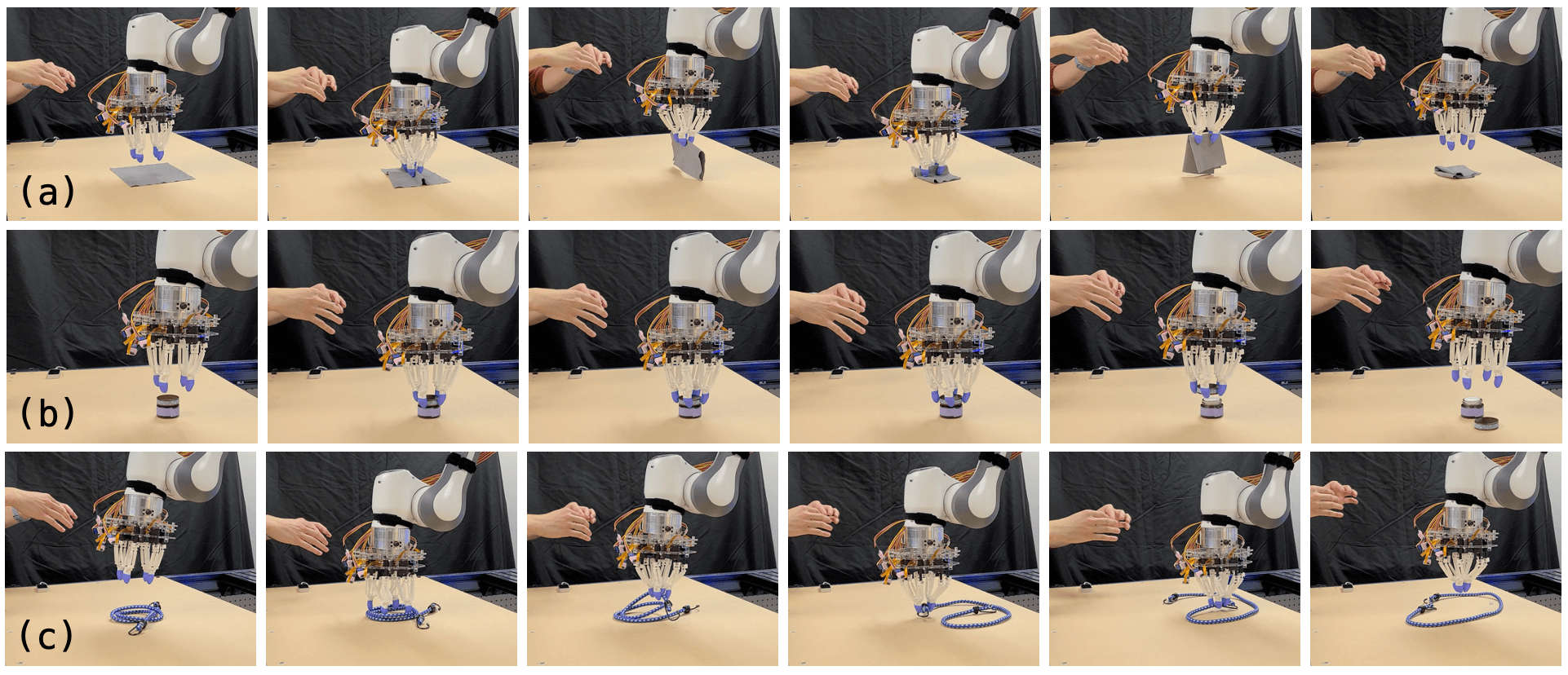}
    \vspace{-1mm}
    \caption{Teleoperation of (a) Cloth folding by folding the cloth twice in two perpendicular directions; (b) Cap opening by twisting the cap and then removing the cap from the bottle; (c) Cable arrangement by loosening the cable loop, and rearranging the cable.}
    \label{fig:teleoperation}
    \vspace{-6mm}
\end{figure*}

\begin{figure}
    \centering
    \includegraphics[width=0.85\linewidth]{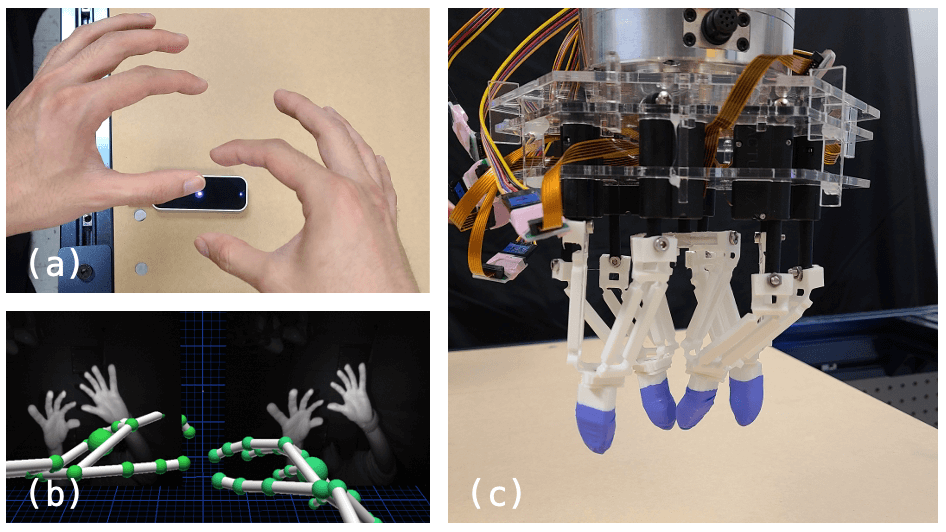}
    \vspace{-1mm}
    \caption{Teleoperation setup with a Leap Motion camera. An operator's hand and finger poses are tracked by the camera and then mapped to the robot arm's and a 9-actuator hand's motions.}
    \label{fig:teleoperation-setup}
    \vspace{-7mm}
\end{figure}

In order to show \name' dexterity, we teleoperate a 9-actuator hand and a Franka robot arm by using a Leap Motion \cite{Ultraleap} camera with its built-in human hand and finger tracking (Fig.~\ref{fig:teleoperation-setup}). We map the operator's left-hand palm translational motions to the Franka robot arm's end-effector (EE) motions and two hands' index and thumb fingers' translational motions to the \name' four-finger motions. We demonstrate three dexterous manipulation tasks: cloth folding, cap opening, and cable rearrangement (Fig.~\ref{fig:teleoperation}). Using real-time teleoperation with human corrections, we can achieve a closed human-in-the-loop control system.

To stabilize and simplify teleoperation, we leverage both \textit{software synergies} in EE space and \textit{hardware synergies} in actuation space. When mapping the operator fingers' motions to \name' EE motions, we pre-define task-specified synergy matrix $S \in \mathbb{R}^{12 \times 12}$ to coordinate finger motions similar to Eigengrasps~\cite{ciocarlie2007dexterous}. Then, the \name' EE poses are converted to a feasible actuation space by using the hand's IK model with projection matrix P from Eq.~\ref{eq:projection}. Thus, given the operator's finger positions as $X = [p_{left\_thumb}, p_{left\_index}, p_{right\_thumb}, p_{right\_index}] \in \mathbb{R}^{12}$, where $p = [x, y, z]$, we map it to the \name' actuation space as $A = P \times IK(S X) \in \mathbb{R}^{9}$.


\subsubsection{Cloth folding}
We fold a cloth in half twice along two perpendicular directions by pinching, lifting, placing, and spreading. The finger motions are aligned in X and Y directions therefore we define the synergy matrix $S_{folding}$ by averaging fingers' positions to a square shape constraint.

\subsubsection{Cap opening}
We open and remove a bottle cap by repeatedly grasping, twisting, and releasing. All fingers move symmetrically around the cap's central point, so we define the synergy matrix $S_{opening}$ by constraining fingers to a circular shape in a polar coordinate frame.

\subsubsection{Cable arrangement}
We straighten a loosely tangled cable by spreading, grasping, and sliding. This can also be realized by aligning the motions to X and Y directions, so we use the folding synergy matrix $S_{cable} = S_{folding}$.

These tasks demonstrate the hand's dexterous motions such as pinching, twisting, and spreading. Benefiting from hand synergies, we can easily and stably control the hand with human teleoperation. The soft materials of the hand allow it to safely and adaptively interact with the environment without damage and compensate for deviations in the kinematics.

\section{Conclusions}
We present~\name, a synergistic dexterous hand framework that utilizes soft Delta robots.~\namespace are low-cost, easy to build and control, and highly configurable regarding design parameters. We characterize the Delta robot's kinematics accuracy, force profile, and workspace range which can guide the design and configuration of our robotic hands. We present two different actuation synergies to reduce the control difficulty and energy consumption while keeping the hands' dexterity. We also show the capability of using our hand framework to further explore various synergistic configurations. We demonstrate diverse object grasping with two hand configurations and quantitative evaluations in the simulator. Furthermore, we teleoperate a 9-actuator hand with a Franka robot arm to complete three dexterous manipulation tasks: cloth folding, cap opening, and cable arrangement.

To extend this work,  we plan to explore task-oriented design optimizations as compared to manual tuning of hand parameters and actuation synergies. In addition, we aim to investigate adaptive control policies for different hand synergies. Finally, instead of closing the control loop through human teleoperation, we would like to add sensors on the hands and fingers to enable automated closed-loop control.
\section*{Acknowledgements}

This work was funded by the NSF under project Grant No.CMMI-2024794. The authors sincerely thank Sarvesh Patil, Moonyoung (Mark) Lee, Sha Yi, Jennifer Yang, and Shashwat Singh for their help in discussions, experiments, and with manuscript revisions.

\bibliography{references}

\end{document}